\documentclass[runningheads]{llncs}
\usepackage[T1]{fontenc}
\usepackage{graphicx}
\usepackage{amsmath}
\usepackage{amssymb}
\usepackage{physics}
\usepackage{algorithm}
\usepackage{algpseudocodex}
\usepackage{tikz}
\usetikzlibrary{arrows.meta,decorations,positioning,shapes}

\newcommand\af{\sigma}
\newcommand\at[2]{\left.#1\right|_{#2}}
\newcommand\inlayer[2]{#1^{\left[#2\right]}}
\newcommand\jacobian{\mathbf{J}}
\newcommand\model{f}

\begin{document}
\title{Towards a perturbation-based explanation for medical AI as differentiable programs}
\author{Takeshi Abe \and Yoshiyuki Asai}
\authorrunning{T. Abe et al.}
\institute{Yamaguchi University}
\maketitle              % typeset the header of the contribution
\begin{abstract}
% The abstract should briefly summarize the contents of the paper in
Recent advancement in machine learning algorithms reaches a point where
medical devices can be equipped with artificial intelligence (AI) models for
diagnostic support and routine automation in clinical settings.
In medicine and healthcare, there is a particular demand for sufficient and
objective explainability of the outcome generated by AI models.
However, AI models are generally considered as black boxes due to their complexity,
and the computational process leading to their response is often opaque.
Although several methods have been proposed to explain the behavior of models by
evaluating the importance of each feature in discrimination and prediction, they
may suffer from biases and opacities arising from the scale and sampling
protocol of the dataset used for training or testing.
To overcome the shortcomings of existing methods, we explore an alternative
approach to provide an objective explanation of AI models that can be defined
independently of the learning process and does not require additional data.
As a preliminary study for this direction of research, this work examines a
numerical availability of the Jacobian matrix of deep learning models that
measures how stably a model responses against small perturbations added to the
input.
The indicator, if available, are calculated from a trained AI model for a given
target input.
This is a first step towards a perturbation-based explanation, which
will assist medical practitioners in understanding and interpreting the response
of the AI model in its clinical application.

\end{abstract}
\section{Introduction}
Recent innovation of machine learning algorithms, including deep learning,
has promoted the use of AI in various scientific and technological fields.
In clinical medicine and healthcare, it is highly expected to introduce AI for
diagnostic support and automation of routine tasks.
However, there are several pressing issues in AI's medical application.
One of them is the fragility of mechanisms to ensure explainability for the
results of classification or prediction task generated by AI models.
This is of practical importance because a stricter standard is imposed on the
responsibility for explaining decisions made by medical practitioners than in
other fields, such as marketing, leading the applications of AI ahead.

In the realm of explainable AI (XAI), there are two typical approaches to
conventional explainability.
One approach is to explain the tendencies of the output from the inherent
characteristics of the modelling or learning algorithm.
For example, there are methods that clarify the contribution of each feature to
prediction or discrimination in generalized linear regression models or decision
tree-based models.
The other approach forms a group of methods to measure the importance of
features independently of the model type, i.e., in a model-agnostic way.
Notable examples of this group include local algorithms for individual targets,
such as LIME~\cite{ribeiro2016} and SHAP~\cite{lundberg2017}.
In both approaches, the indices of explainability are calculated based on the
accuracy or other performance measures on correct responses.
It is also assumed that the dataset used for learning and/or testing follows
the distribution of the target population.
There is a risk of bias when this premise fails to hold.
Moreover, there is opacity when information about the dataset in use cannot be
accessed~\cite{drukker2023}.
Some criticism of current XAI technology argues that it is more appropriate to
use models designed to be interpretable from the start, rather than trying to
explain from black-boxed models post-hoc~\cite{rudin2019}.
However, methodological significance resides in methods improving explainability
of the AI models actually utilized in clinical practice.

In this work, focusing on deep learning models, we explore an alternative
approach to explain how sensitive the model's response is with respect to
a perturbation to each instance.
Here an instance means an input subject of the model, and perturbation refers to
a small change in the input features of an instance.
We aim at providing this approach with the following desirable properties.
(a) It can be applied to trained AI models and does not depend on the learning
process of them.
(b) It is free from additional data collection.
(c) It measures the stability of the output against the virtual variation of
input features.
(d) It returns a value for each instance, not the average value for the entire
dataset, making it local.
In addition, (e) it offers explainability not only for a single feature but also
for interactions between features.
We call such an approach a perturbation-based explanation (PBX).
Complementing existing indicators for XAI with a PBX will facilitate
medical practitioners to appropriately understand and interpret the behavior of
a medical AI.

By considering the AI model as a function with argument of many features, the
impact of perturbations on each feature is often represented as a gradient.
From the components of the partial derivative at the point of the instance, it
can be judged how stably each feature is outputting against perturbations.
Examples of such gradient-based approaches include SmoothGrad~\cite{smilkov2017}
and Integrated Gradients~\cite{sundararajan2017}, but the unification with
perturbation-based approaches is still under development.

For a preliminary study of a PBX, this work examines when and how a partial
derivative of the deep learning model, as a multivariate function, is
computationally available.
In particular, since the Jacobian matrix is the most fundamental kind of partial
derivatives, we concentrate on the numerical computability of the model's
Jacobian matrix in this work.
To formalize our argument for general applicability, the next Methods section
describes the standard model of deep learning, representing a multivariate
function.
Then we demonstrate that the Jacobian matrix of this function can be calculated
by a certain forward pass algorithm in the Results section.
We summarizes the implication of this work and potential future development in
the Discussion and Conclusion sections.

\section{Methods}
To represent the standard model of deep learning, we employ the following
notation adapted from Higham et al.~\cite{higham2019}.
\begin{figure}
\centering
\begin{tikzpicture}[
neuron/.style = {draw,circle},
right triangle/.style =
{regular polygon,regular polygon sides=3,shape border rotate=270,minimum size=0.3cm}
]

\node[] (i1) at (-1,4) {$x_1$};
\node[] (i2) at (-1,3) {$x_2$};
\node[] (i3) at (-1,1) {$x_m$};
\node[] (y) at (-1,0.4) {$x$};

\node[] (i1) at (-0.5,4) {$\rightarrow$};
\node[] (i2) at (-0.5,3) {$\rightarrow$};
\node[] (i3) at (-0.5,1) {$\rightarrow$};

\node[neuron,label=layer $1$] (l1-n1) at (0,4) {};
\node[neuron] (l1-n2) at (0,3) {};
\node[] at (0,2) {$\vdots$};
\node[neuron] (l1-n3) at (0,1) {};

\node[neuron,label=layer 2] (l2-n1) at (2,4) {};
\node[neuron] (l2-n2) at (2,3) {};
\node[] at (2,2) {$\vdots$};
\node[neuron] (l2-n3) at (2,1) {};

\node[neuron,label=layer 3] (l3-n1) at (4,4) {};
\node[neuron] (l3-n2) at (4,3) {};
\node[] at (4,2) {$\vdots$};
\node[neuron] (l3-n3) at (4,1) {};

\node[neuron,label=layer $L-1$] (lL-1-n1) at (6,4) {};
\node[neuron] (lL-1-n2) at (6,3) {};
\node[] at (6,2) {$\vdots$};
\node[neuron] (lL-1-n3) at (6,1) {};

\node[neuron,label=layer $L$] (lL-n1) at (8,4) {};
\node[neuron] (lL-n2) at (8,3) {};
\node[] at (8,2) {$\vdots$};
\node[neuron] (lL-n3) at (8,1) {};

\node[] (o1) at (8.5,4) {$\rightarrow$};
\node[] (o2) at (8.5,3) {$\rightarrow$};
\node[] (o3) at (8.5,1) {$\rightarrow$};

\node[] (y1) at (9,4) {$y_1$};
\node[] (y2) at (9,3) {$y_2$};
\node[] (y3) at (9,1) {$y_n$};
\node[] (y) at (9,0.4) {$y$};

\node[] (a1) at (0,0.5) {$= a^{[1]}$};
\node[] (a2) at (2,0.5) {$a^{[2]}$};
\node[] (a3) at (4,0.5) {$a^{[3]}$};
\node[] (aL-1) at (6,0.5) {$a^{[L-1]}$};
\node[] (aL) at (8,0.5) {$a^{[L]} =$};

\foreach \m in {1,2,3} {
  \foreach \n in {1,2,3} {
    \draw (l1-n\m) -- (l2-n\n);
    \draw (l2-n\m) -- (l3-n\n);
    \draw (l3-n\m) -- (lL-1-n\n);
    \draw (lL-1-n\m) -- (lL-n\n);
  }
}

\node[fill=white,minimum width=1cm,minimum height=2.5cm] at (5,2.5) {};

\node[fill=white,minimum width=0.5cm,minimum height=1.5cm] at (-1,2) {$\vdots$};
\node[fill=white,minimum width=0.5cm,minimum height=1.5cm] at (1,2) {$\vdots$};
\node[fill=white,minimum width=0.5cm,minimum height=1.5cm] at (3,2) {$\vdots$};
\node[fill=white,minimum width=0.5cm,minimum height=1.5cm] at (5,2) {$\vdots$};
\node[fill=white,minimum width=0.5cm,minimum height=1.5cm] at (7,2) {$\vdots$};
\node[fill=white,minimum width=0.5cm,minimum height=1.5cm] at (9,2) {$\vdots$};

\node[fill=white] at (5,4) {$\cdots$};
\node[fill=white] at (5,3) {$\cdots$};
\node[fill=white] at (5,1) {$\cdots$};

\node[fill=white] (Wa2) at (1,2.5) {$W^{[2]}; \sigma^{[2]}$};
\node[fill=white] (Wa3) at (3,2.5) {$W^{[3]}; \sigma^{[3]}$};
\node[fill=white] (WaL) at (7,2.5) {$W^{[L]}; \sigma^{[L]}$};

\end{tikzpicture}
\caption{Schematic diagram for the standard model of deep learning.}\label{fig:model}
\end{figure}
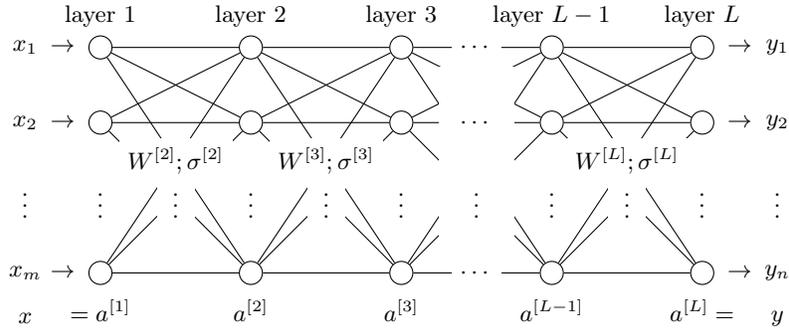
Figure~\ref{fig:model} illustrates the layered structure of a model.
\begin{itemize}
\item $L$:
the number of layers in the model

\item $\inlayer{n}{l}$:
the number of nodes in $l$-th layer ($l = 1, \dots, L$)

\item $x$:
the column vector of input (of length $m = \inlayer{n}{1}$)
\[x = \inlayer{a}{1} \in \mathbb{R}^m\]

\item $\inlayer{W}{l}$:
weight matrix for weighted input at $l$-th layer ($l = 2, \dots, L$):
\[\inlayer{W}{l} = \left(\inlayer{w}{l}_{i j} \right)_{i,j}: \inlayer{n}{l} \times \inlayer{n}{l-1} \text{ matrix}\]

\item $\inlayer{z}{l}$:
weighted input at $l$-th layer ($l = 2, \dots, L$):
\[\inlayer{z}{l} = \inlayer{W}{l} \inlayer{a}{l-1}\]

\item $\inlayer{\af}{l}$:
activation function at $l$-th layer ($l = 2, \dots, L$):
\[\inlayer{\af}{l}: \mathbb{R}^{\inlayer{n}{l}} \rightarrow \mathbb{R}^{\inlayer{n}{l}}\]

\item $\inlayer{a}{l}$:
activation at $l$-th layer ($l = 1, \dots, L$):
\[\inlayer{a}{l} = \left\{\begin{array}{ll}
  x & \text{ if } l = 1;\\
  \inlayer{\af}{l}(\inlayer{z}{l}) & \text{ otherwise}.
\end{array}\right.\]

\item $y$:
the column vector of output (of length $n = \inlayer{n}{L}$):
\[y = \inlayer{a}{L} \in \mathbb{R}^n\]
\end{itemize}
Note that, unlike \cite{higham2019}, we exclude the so-called bias term in
this representation because it can be replaced with additional elements of
weight matrices by introducing an corresponding input variables that are always
constant 1.
Another modification to the formulation of \cite{higham2019} is that
the activation function $\inlayer{\af}{\cdot}$ can differ from layer to layer.
Moreover, $\inlayer{\af}{\cdot}$ is allowed to be multivariate.

In this work we suppose that every element of vectors and matrices is
real-valued in a given model.
Hereafter the model is denoted as a vector-valued multivariate function
$\model: \mathbb{R}^m \rightarrow \mathbb{R}^n$: $y = \model(x).$
That is, $y_i = \model_i(x)$ for $i = 1, \dots, m$, if written in an element-wise way.

\section{Results}
The following theorem states a sufficient condition that the model's Jacobian
matrix can be computed.
\begin{theorem}\label{thm:1}
Let be $\model$ a model represented in the above notation.
if $\inlayer{\af}{l}$ has a Jacobian matrix that is computable at any point
for each layer $l = 2, \dots, L$, then $\model$'s Jacobian matrix $\jacobian_\model$
is computable at a given point.
\end{theorem}
\begin{proof}
Since $y = \inlayer{\af}{L}(\inlayer{W}{L}
\inlayer{\af}{L-1}(\inlayer{W}{L-1} \cdots
\inlayer{\af}{2}(\inlayer{W}{2} x)
\cdots) )$, the matrix version of the chain rule tells that $\model$'s Jacobian
matrix is decomposed into \(\inlayer{\af}{l}(\inlayer{W}{l} \cdot)\)'s Jacobian
matrices:
\begin{equation}\label{eq:jacobian}
\jacobian_\model
=
\pdv{y}{\inlayer{a}{L-1}}
\pdv{\inlayer{a}{L-1}}{\inlayer{a}{L-2}}
\cdots
\pdv{\inlayer{a}{2}}{x}
=
\prod_{l=2}^L
\pdv{\inlayer{a}{l}}{\inlayer{a}{l-1}},
\end{equation}
The multiplication in the above equation means the matrix product.
In addition, each element of \(\pdv{\inlayer{a}{l}}{\inlayer{a}{l-1}}\) is as follows:
\begin{equation}
\pdv{\inlayer{a}{l}_i}{\inlayer{a}{l-1}_j}
= \grad{\inlayer{\af}{l}_i(\inlayer{z}{l})} \bullet \left(\begin{array}{c}
  \pdv{\inlayer{z}{l}_1}{\inlayer{a}{l-1}_j}\\
  \vdots\\
  \pdv{\inlayer{z}{l}_{\inlayer{n}{l}}}{\inlayer{a}{l-1}_j}
\end{array}\right)
= \grad{\inlayer{\af}{l}_i(\inlayer{z}{l})} \bullet \left(\begin{array}{c}
  \inlayer{w}{l}_{1 j}\\
  \vdots\\
  \inlayer{w}{l}_{\inlayer{n}{l} j}\\
\end{array}\right)
\end{equation}
where \(\bullet\) is the dot product and
\(\grad{\inlayer{\af}{l}_i(\inlayer{z}{l})}\) denotes the gradient of the $i$-th
node's activation at layer $l$, which is computable because its transpose
corresponds to a row vector of \(\inlayer{\af}{l}\)'s Jacobian matrix.
\qed
\end{proof}

While Theorem~\ref{thm:1} looks similar with a well-known property of standard
deep learning models that the backpropagation algorithm takes advantage of,
it is more general as it can hold even for activation functions that do not meet
the element-wise diagonal condition~\cite{higham2019}.

From Theorem~\ref{thm:1}'s proof in the above, we extract a foward-pass
algorithm to calculate the Jacobian matrix at a given input numerically
(Algorithm~\ref{alg:1}).
\begin{algorithm}[h]
\caption{}\label{alg:1}
\begin{algorithmic}[1]
\Require Input: a given model $\model$; an input $x \in \mathbb{R}^m$ of model $\model$.
\Statex Procedures to calculate \(\inlayer{\af}{l}\)'s Jacobian matrix
\(\jacobian_{\inlayer{\af}{l}}\) at \(z\) \((l = 2, \dots, L)\);
\Statex \(\at{\jacobian_{\inlayer{\af}{l}}}{z} =
\left(\begin{array}{c}
  (\grad{\inlayer{\af}{l}_1(z)})^\top\\
  \vdots\\
  (\grad{\inlayer{\af}{l}_{\inlayer{n}{l}}(z)})^\top
\end{array}\right)\)
\Ensure Output: $\model$'s Jacobian matrix $\jacobian_\model$ at $x$;
\Statex $\at{\jacobian_\model}{x} =
\at{\left(\frac{\partial \model_i}{\partial x_j}\right)_{i,j}}{x}
\quad (i = 1, \dots, n; j = 1, \dots, m)$.
\State $\inlayer{a}{1} \gets x$
\State $\inlayer{J}{1} \gets \operatorname{I}_m$: the identity matrix of size \(m\)
\For{$l = 2$ to $L$}
\State $\inlayer{z}{l} \gets \inlayer{W}{l} \inlayer{a}{l-1}$
\State $\inlayer{a}{l} \gets \inlayer{\af}{l}(\inlayer{z}{l})$
\State $\inlayer{J}{l} \gets \at{\jacobian_{\inlayer{\af}{l}}}{\inlayer{z}{l}}
\inlayer{W}{l} \inlayer{J}{l-1}$
\EndFor
\State \Return $\inlayer{J}{L}$
\end{algorithmic}
\end{algorithm}

\section{Discussion}
The numerical computation of the Jacobian matrix of a deep learning model
demonstrated in our results can be applied to a wide variety of pre-trained deep
learning models potentially used for classification and regression tasks in
a medical AI.
Our treatment of activation functions is so general that it covers not only
the usual activation functions of a scalar argument, e.g., the logistic or
softplus function, but also the one taking the argument of two or more nodes,
such as the softmax function often appended in the output layer.
When applied, this local method calculates for each individual input data, so it
does not require additional data.
Since Algorithm~\ref{alg:1} is implemented as a simple extension of the forward
differentiation algorithm, it is straightforward to port the algorithm to any
libraries of recent real-world machine learning frameworks, such as PyTorch and
TensorFlow.

The proposed algorithm calculates the exact value of the Jacobian matrix at given
instance, provided that the computation of the gradient of activation functions
is exact.
This fact makes a difference when comparing with the numerical calculation of the
Jacobian matrix by the classical finite difference method~\cite{press2007}.
Also, the proposed algorithm will outperform in terms of computational efficiency,
as it evaluates the target multivariate function just in one pass, while
the finite difference method evaluates the model \(m+1\) times, i.e.,
one more times than the number of input features.
Another advantage of Algorithm~\ref{alg:1} is that it generates the Jacobian
matrix \(\inlayer{J}{l}\) of any initial part of the model up to an intermediate
layer \(l\) as a byproduct.
These matrices may be as important as the full model's e.g. when investigating
how the encoder part of an autoencoder model is sensitive to a small
perturbation to the input.

On the other hand, there are some limitations to the proposed method.
One of them is the assumption that the activation function of each
layer must have a (computable) partial derivative at an arbitrary point.
For instance, the ReLU function, a popular choice of the activation
function in many AI applications, has a singular point at zero.
When the proposed algorithm is applied to a deep learning model containing such
a piecewise differentiable activation function, involving its Jacobian matrix at
singular points may produce unexpected results.
Another limitation is that the proposed algorithm does not specify how to
prepare the procedure to compute the Jacobian matrix or gradient of activation
functions in the trained model.
While the information necessary for coding the prodecure is available in the
model, this work leaves the details on how to extract and transform them into a
computable function.

Regarded as an indicator of the stability or sensitivity of a model's
response to perturbed inputs, the computed gradient shows which input feature
is the most influential per unit change.
One compares the columns of the gradient for the purpose.
On the other hand, comparing the rows of a computed Jacobian matrix finds
which element of responses is the most affected for the same perturbation.
This type of knowlege helps especially when the each element of responses is
of the same unit, e.g., representing the probability that the input target
belongs to a different category posed in a classification task.

\section{Conclusion}
This work is our first step towards a PBX that makes an objective interpretation
of medical AI models without the burden of data cost.
The Jacobian matrix, once computed, favorably satisfies PBX's properties (a) to (d)
mentioned in the Introduction section.
However, to fulfill (e), the matrix is insufficient for revealing the stability
of responses around given perturbed input since it lacks the information of
interactions among input variables.
That is, higher-order partial derivatives are called for when adding
perturbations to multiple input features at the same time.

For future development of PBX, it is promising to utilize a wider range of
mathematical tools, including the concept of perturbation theory developed in
the analysis of mathematical models succeeded in physics and engineering.
Moreover, some extension of this work is required for models built with machine
learning algorithms other than deep learning.

\bibliographystyle{splncs04}
\bibliography{icann2024}

\begin{thebibliography}{1}
\providecommand{\url}[1]{\texttt{#1}}
\providecommand{\urlprefix}{URL }
\providecommand{\doi}[1]{https://doi.org/#1}

\bibitem{drukker2023}
Drukker, K., Chen, W., Gichoya, J.W., Gruszauskas, N.P., {Kalpathy-Cramer}, J.,
  Koyejo, S., Myers, K.J., S{\'a}, R.C., Sahiner, B., Whitney, H.M., Zhang, Z.,
  Giger, M.L.: Toward fairness in artificial intelligence for medical image
  analysis: Identification and mitigation of potential biases in the roadmap
  from data collection to model deployment. Journal of Medical Imaging
  \textbf{10}(6),  061104 (Apr 2023). \doi{10.1117/1.JMI.10.6.061104}

\bibitem{higham2019}
Higham, C.F., Higham, D.J.: Deep {{Learning}}: {{An Introduction}} for
  {{Applied Mathematicians}}. SIAM Review  \textbf{61}(4),  860--891 (Jan
  2019). \doi{10.1137/18M1165748}

\bibitem{lundberg2017}
Lundberg, S.M., Lee, S.I.: A {{Unified Approach}} to {{Interpreting Model
  Predictions}}. In: Advances in {{Neural Information Processing Systems}}.
  vol.~30. Curran Associates, Inc. (2017)

\bibitem{press2007}
Press, W.H., Teukolsky, S.A., Vetterling, W.T., Flannery, B.P.: Numerical
  {{Recipes}} 3rd {{Edition}}: {{The Art}} of {{Scientific Computing}}.
  Cambridge University Press, USA, 3 edn. (Aug 2007)

\bibitem{ribeiro2016}
Ribeiro, M.T., Singh, S., Guestrin, C.: "{{Why Should I Trust You}}?":
  {{Explaining}} the {{Predictions}} of {{Any Classifier}}. In: Proceedings of
  the 22nd {{ACM SIGKDD International Conference}} on {{Knowledge Discovery}}
  and {{Data Mining}}. pp. 1135--1144. ACM, San Francisco California USA (Aug
  2016). \doi{10.1145/2939672.2939778}

\bibitem{rudin2019}
Rudin, C.: Stop explaining black box machine learning models for high stakes
  decisions and use interpretable models instead. Nature Machine Intelligence
  \textbf{1}(5),  206--215 (May 2019). \doi{10.1038/s42256-019-0048-x}

\bibitem{smilkov2017}
Smilkov, D., Thorat, N., Kim, B., Vi{\'e}gas, F., Wattenberg, M.:
  {{SmoothGrad}}: Removing noise by adding noise (Jun 2017).
  \doi{10.48550/arXiv.1706.03825}

\bibitem{sundararajan2017}
Sundararajan, M., Taly, A., Yan, Q.: Axiomatic attribution for deep networks.
  In: Proceedings of the 34th {{International Conference}} on {{Machine
  Learning}} - {{Volume}} 70. pp. 3319--3328. {{ICML}}'17, JMLR.org, Sydney,
  NSW, Australia (Aug 2017)

\end{thebibliography}
\end{document}